# Few-shot learning for automated content analysis: Efficient coding of arguments and claims in the debate on arms deliveries to Ukraine

Dr. Jonas Rieger [1], j.rieger@leibniz-hbi.de
Dr. Kostiantyn Yanchenko [2], kostiantyn.yanchenko@uni-hamburg.de
Mattes Ruckdeschel [1], m.ruckdeschel@leibniz-hbi.de
Dr. Gerret von Nordheim [2], gerret.vonnordheim@uni-hamburg.de
Prof. Dr. Katharina Kleinen-von Königslöw [2], katharina.kleinen@uni-hamburg.de
Dr. Gregor Wiedemann [1], g.wiedemann@leibniz-hbi.de

[1] Leibniz Institute for Media Research | Hans-Bredow-Institut (HBI)
Rothenbaumchaussee 36, 20148 Hamburg

[2] Fachgebiet Journalistik/Kommunikationswissenschaft, Universität Hamburg
Von-Melle-Park 5, 20146 Hamburg

**Abstract**

Pre-trained language models (PLM) based on transformer neural networks developed in the field of natural language processing (NLP) offer great opportunities to improve automatic content analysis in communication science, especially for the coding of complex semantic categories in large datasets via supervised machine learning. However, three characteristics so far impeded the widespread adoption of the methods in the applying disciplines: the dominance of English language models in NLP research, the necessary computing resources, and the effort required to produce training data to fine-tune PLMs. In this study, we address these challenges by using a multilingual transformer model in combination with the adapter extension to transformers, and few-shot learning methods. We test our approach on a realistic use case from communication science to automatically detect claims and arguments together with their stance in the German news debate on arms deliveries to Ukraine. In three experiments, we evaluate (1) data preprocessing strategies and model variants for this task, (2) the performance of different few-shot learning methods, and (3) how well the best setup performs on varying training set sizes in terms of validity, reliability, replicability and reproducibility of the results. We find that our proposed combination of transformer adapters with pattern exploiting training provides a parameter-efficient and easily shareable alternative to fully fine-tuning PLMs. It performs on par in terms of validity, while overall, provides better properties for application in communication studies. The results also show that pre-fine-tuning for a task on a near-domain dataset leads to substantial improvement, in particular in the few-shot setting. Further, the results indicate that it is useful to bias the dataset away from the viewpoints of specific prominent individuals.

pre-trained language models; transformer adapters; claim, argument and stance detection; automatic media content analysis; quality metrics

# 1 Introduction

The advent of pre-trained language models based on transformer neural networks such as BERT (Devlin et al., 2019) has revolutionized the field of natural language processing (NLP). This novel approach has the potential to overcome the limitations of previous computational text research by effectively capturing complex semantic concepts with respect to the sequentiality and contextuality of natural language (Wiedemann & Fedtke, 2021). The successful transfer of linguistic and semantic knowledge from self-supervised pretraining with a so-called masked language model task, i.e., a cloze test for the neural network to predict randomly masked words in a given sentence, to any target task in NLP has fueled the hope that powerful text classifiers can also be built for automatic content analysis with comparatively little training data. This task is researched in NLP under the term *few-shot learning* (FSL). So far, the implementation of these new methods in communication science has been somewhat limited. The lack of accessible and efficient ways to fine-tune large language models on the practical-technical side and the lack of best practice cases on the methodological side seem to hinder the adoption and development of new standards. Both problems are interrelated and can only be solved together.

This paper introduces a new implementation that we believe is particularly suitable for the widespread adoption of transfer learning in communication studies: a parameter-efficient fine-tuning (PEFT) based on adapters in combination with semi-supervised pattern-exploiting training (PET). The adapter approach (Pfeiffer et al., 2020) allows for the fine-tuning of transformer-based classifiers more efficiently in terms of training time and model size. The PET approach (Schick & Schütze, 2021) allows for the significant improvement of classification performance on small training datasets by using not only labeled texts but also the semantic knowledge from the labels themselves to learn from. We demonstrate the potential of our approach by applying it to a prototypical communication studies problem — the identification of claims and arguments in the news media debate.

Claims are a subject of interest for communication scholars because they serve as the fundamental conveyors of a perspective, or stance, and can therefore be utilized for a systematic examination of balance in media reporting (de Bruycker & Beyers, 2015). Arguments, in turn, are important because they reveal the underlying reasoning and evidence used to support a claim, which can aid in the evaluation of the validity and strength of a given perspective presented in a media debate (Meyers et al., 2000). Past studies on argument mining have indicated that identifying claims and arguments following a rigid formal structure such as in legal documents or student essays (Daxenberger et al., 2017; Hüning et al., 2022) is considerably easier compared to the more complex and latent argumentative structures found in news media debates. The goal of this paper is to show how traditional content analytic approaches to identify claims and arguments can be aided by state-of-the-art methods from supervised learning with a special focus on the promises and limitations of FSL. Along with this, we address the associated validity, reliability, reproducibility, and replicability concerns.

The proposed approach is demonstrated through a case study of the *Waffenlieferung* (arms delivery) debate in Germany. The issue of potential arms deliveries to Ukraine gained prominence in German news media several weeks before Russia's full-scale invasion of Ukraine and has since remained a recurring topic (Maurer et al., 2023). Over time, the *Waffenlieferung* issue proved to be both sensitive and complex and provoked various meta-debates, such as the alleged one-sided coverage of the issue by the German mainstream media (Precht & Welzer, 2022). At the same time, just as many other highly polarized issues such as vaccination or xenophobia, the *Waffenlieferung* debate implies one of the two opposing viewpoints — either to supply arms or not. From this perspective, the task of detecting claims and arguments on this topic can be considered basic compared to other more nuanced and multi-polar debates. Therefore, our case serves as an ideal starting point for exploring the capabilities of language models in identifying claims and arguments in news media. The straightforward polarity of the arms delivery debate allows us to focus on the core competencies of these models before future research may venture into more nuanced and intricate discourses.

At a broader level, our study also aims to overcome the three barriers to the more widespread adoption of computational methods in communication science identified by Baden et al. (2022): a) Using a real-life communication research use-case allows us to focus our method development from the outset on the requirements of measurement validity; b) moving beyond previous studies using automated analysis to classify either entire documents or identify specific expressions within texts, we capture arguments, claims, and stances as well as the relationships between them on a sentence level, allowing us to understand the argumentative complexity of a text; c) by working with multi-lingual transformer models the usefulness and applicability of our method developments are not limited to English language texts.

The rest of the contribution is structured as follows: Section 2 reviews the concepts of argument, claim, and FSL; Section 3 outlines our approach to automated identification of arguments and claims, including data, codebook, and language models; Section 4 presents our findings; and Section 5 discusses the implications of our study, including the key insights for communication scholars, limitations, and future research directions. Analysis scripts, models, and results as well as further technical details can be accessed at the corresponding OSF repository[1].

---

[1] https://osf.io/tayxq/?view_only=7a8771ba39534782877bc83e0eb09944

## 2 Related work

The findings presented in this paper build upon a couple of research fields covering the conceptual definition of claim and argument, argument mining with the facet of stance detection, and FSL for text classification.

### 2.1 Claims and arguments

The concepts of claim and argument are prominent in NLP and machine learning fields where argument mining has been an important classification task since the beginning of the 2010s (Cabrio & Villata, 2018; Lippi & Torroni, 2016). At the same time, these concepts are less prevalent among communication scholars despite their relevance to many seminal topics in communication research. One such topic is the viewpoint diversity of the media discourse which is commonly approached from the perspective of article-level frames (Baden & Springer, 2017; Masini & Aelst, 2017) but can also be studied at the sentence level (Voakes et al., 1996), with each unique argument representing a different viewpoint or micro-frame. Another example would be the relevance of (unsupported) claims and (misleading) arguments for disinformation research (Vlachos & Riedel, 2014; Cook et al., 2017).

Conceptually, both claims and arguments originate from Argumentation Theory and can be operationalized differently depending on the research domain and school of thought one subscribes to (for a comprehensive overview of various theoretical approaches to the topic, see van Eemeren et al., 2014). As it is the goal in this study to automatically detect claims and arguments drawing on the common-sense perspective of a media text's reader, we will further focus on the basic claim-premise model that underlies most existing definitions of an argument (Goodman, 2018, p. 593) and has long been accepted in NLP (Walton, 2009).

Table 1. The examples of claims and arguments regarding arms deliveries to Ukraine from the German media debate (claims are given in italics, premises are in regular font)

| Label | Example |
| --- | --- |
| claim for | *The strongest demands for arms deliveries come from the CDU.* |
| claim against | *Saxony's Prime Minister Michael Kretschmer (CDU) has spoken out vehemently against the delivery of heavy weapons to Ukraine.* |
| argument for | But if you really want to help the Ukrainians stop Putin's butchery troops and drive them out of the country, *you must now provide them with weapons* to go on the offensive. |
| argument against | This is due to the federal government not wanting to export armaments to crisis areas, and because of its past, *Berlin does not want to rush forward with arms deliveries* to the former Soviet Union. |

According to the claim-premise model, a *claim* – sometimes also referred to as *conclusion*, *standpoint*, *thesis*, or *proposition* – is an unjustified assertion about reality or, as Biran and Rambow put it, an "utterance which conveys subjective information" (2011, p. 364). When considered in the context of a polarized debate implying two opposing viewpoints, a claim can be classified by its stance as a claim *for* or *against* certain actions, policies, etc. Such a type of claim is called prescriptive (for the typology of claims, see Eemeren et al., 2014, p. 14). A claim is a self-sufficient unit that can be studied independently but is also a key component of an argument. To be considered an argument, a claim needs to be supported by one or several *premises*, sometimes also called *data*, *evidence*, or *reasons* (Lippi & Torroni, 2016). Premises are statements – or more generally pieces of information – that provide justification for an otherwise unsupported claim (Hüning et al., 2022). Thus, an argument can be viewed as a "set of statements, one of which [the claim] is meant to be supported by the other(s) [the premise(s)]" (Munson et al., 2004, p. 5). Often, arguments can be identified by so-called *argument markers* (Eckle-Kohler et al., 2015) – specific signal words and expressions associated with reasoning, e.g., *because*, *therefore*, *since*, etc. However, in natural language, premises can also connect with claims without argument markers resulting in less manifest argumentative structures. Table 1 provides illustrations of claims for/against, as well as arguments for/against in the context of our exemplary study case.

Two issues should be noted about the claim-premise model. First, the presented operationalization is deliberately minimalistic to match our practical interest in claims and arguments in news media discourse (for more extended versions of the claim-premise model, see Toulmin, 2003). Second, we acknowledge that argumentative structures are field- and context-dependent which means that the same utterances can take different functional roles depending on their position and meaning within the given text (Eemeren et al., 2014). To account for this, we incorporate contextual information when identifying claims and arguments both at the stages of coding (cf., Section 3.2) and model training (cf., Section 3.4).

## 2.2 Few-shot learning based on pre-trained language models

For the automated classification of claims and arguments given a set of labeled data, several large pre-trained language models (PLM) have been published in NLP such as RoBERTa (Liu et al., 2019) for English texts, as well as other PLMs specialized for different languages. For the processing of *arbitrary* languages or multilingual texts, the use of XLM-RoBERTa (Conneau et al., 2020) is well suited. This offers a possibility to incorporate existing argumentative datasets in foreign languages as near-domain pre-training (Toledo-Ronen et al., 2020).

The common way of applying these models is standard fine-tuning on the corresponding downstream task (Devlin et al., 2019). By default, all parameters of the PLM are updated in a rather time- and resource-consuming process of gradient descent using the backpropagation algorithm. For this, fine-tuning requires a modern graphics processing unit (GPU) with sufficient memory. Later usage of the fine-tuned PLM is only possible by saving the entire model — for BERT and its successors a size of several gigabytes. Both characteristics may hinder the adaptation of the technology in applied science fields, where such hardware resources are often not readily available. To overcome these disadvantages, so-called adapter transformers have been proposed in NLP research (Pfeiffer et al., 2020). For adapters, model parameters of an initial PLM are frozen, and training is only performed on a set of additional parameter layers that are interwoven with the initial model. The first proposals for these additional layers were limited to bottleneck adapters, which mainly consist of two feed-forward neural nets with a down- and an up-projection. Several studies were able to confirm the authors' finding that this architecture performs on par with standard fine-tuning and mitigates negative effects such as the 'catastrophic forgetting' of pretrained knowledge (He et al., 2021).

For real-world applications such as automatic content analysis in communication science, adapters have two major advantages over standard fine-tuning. First, training time and model size can be reduced drastically because only the adapter module part of the entire model needs to be trained (cf., Table 7). Consequently, only the additionally trained parameters need to be stored to ensure model reproducibility. Due to their much smaller size, archiving, sharing, and reuse of trained models is greatly facilitated. Second, due to the regularization effect of freezing the initial PLM parameters, adapter training is less prone to overfitting than standard fine-tuning. This means that during training, for example, the number of epochs[2] can be chosen more liberally and does not require time-consuming tuning. It also means that performance results in repeated runs vary to a lesser extent due to random weight initializations than for standard fine-tuning.

---

[2] The number of epochs refers to how often the learning algorithm iterates over the entire dataset in training.

Another common problem of real-world applications of supervised learning is the limited amount of training data. A special scenario for this, referred to as FSL in NLP, is pattern-exploiting training (PET; Schick & Schütze, 2021). The basic idea is that a model should not only learn from coded training data but also from instructions such as label names and label definitions to perform a classification task. In this scenario, a training example is injected together with distinct label patterns from the category codes into a so-called verbalizer statement, before being presented to the model. Instead of the prediction of pure (context-free) class labels (aka category labels for communication scientists), PET predicts masked tokens in the language modeling objective within predefined patterns (cf., Section 3.2). By this, it yields the pattern most likely filling a designated slot in the verbalizer that also contains the given training example. The authors were able to show that this approach also works well in realistic few-shot settings where prompt engineering and hyper-parameter tuning is not possible due to limited training data (Schick & Schütze, 2022). However, PET requires some human effort to define good pattern-verbalizer pairs.

## 3 Methodology

In our study, we demonstrate the potential of sequence classification using PLMs in the context of FSL for automated content analysis. For this, we develop a new approach to FSL combining two ideas from the NLP literature: adapters and PET. We compare our approach against other FSL methods (cf., Table 5) concerning their validity, reliability, replicability, reproducibility, and feasibility (cf., Table 2) for the task of predicting argumentative sentences (claim/argument) and their stance (for/against) on a dataset of newspaper articles on the topic of German arms deliveries to Ukraine.

Table 2. Definition of quality criteria in automated content analysis.

| Term | Description | Metric(s) |
| --- | --- | --- |
| Validity | To what extent are the results consistent with a gold standard (here: human coder label)? | Accuracy, precision, recall, (macro-) F1 |
| Reliability* | To what extent do repeated runs of the same methods on the same data produce similar results? | Deviation of validity metrics |
| Replicability* | To what extent does applying the same method to different (related) data produce similar results? | Deviation of validity metrics |
| Robustness* | To what extent do different parameter choices for the same model lead to different results? | Deviation of validity metrics |
| Reproducibility | Given the same model and data, (under what conditions) is it possible to obtain identical results? | Equality check, model size |
| Feasibility | Which resources are required for modeling? | Computation time, resource usage |

*Note.* *We refer to these terms regarding validity assessment, while it is also possible to define them standalone on the predictions themselves. We do not investigate robustness in our experiments.

**3.1 Data**

The initial sample from the German media debate on arms deliveries to Ukraine included all articles mentioning *Ukraine* and *Waffen* [weapons] published between 1 January 2022 and 30 November 2022 in 145 German media outlets, a total of 26,057 articles (translation of German search terms in squared brackets). The corpus comes from the digital archive of the news magazine *Der Spiegel* and was originally compiled for a data journalism project. Within this initial corpus, we conducted additional searches using the combination of *Waffen* AND *L-/liefer* [deliver*] (in one sentence) OR militärisch* [military] & U-/unterstütz* [support] (in one sentence) and retrieved 14,697 articles for our closed corpus. To validate the search string (Mahl et al., 2022), 200 articles were randomly sampled from the initial corpus and labeled by two coders as either relevant or irrelevant to the topic of arms deliveries from Germany to Ukraine (Krippendorff's α = 0.902). Comparing the labels from manual coding with the labels generated by applying the search string revealed a precision score of 0.79 and a recall of 0.86, which were deemed adequate for our task. In a final step, the closed corpus was filtered further to include only leading German-language newspapers, both daily and weekly. This final corpus contained 7,301 articles from 22 media outlets (see Appendix 1 of the Supplementary Information file in the OSF) and was used to code claims and arguments.

## 3.2 Codebook

The codebook explicated how to systematically code arguments for/against and claims for/against arms deliveries to Ukraine, as defined on the previous pages. The four mutually exclusive codes were to be assigned at the sentence level. Both direct and reported claims/arguments were coded. Thus, sentences like "We must deliver arms now! — said X" and "X called for immediate arms supplies" were treated identically. Sentences that merely described the fact of arms deliveries and did not provide clues regarding the stance of an actor X on the topic were not coded, e.g.: "Germany can deliver 100 rocket-propelled grenades to Ukraine" (here, capability rather than a stance).

When coding for claims and arguments, coders were urged to always consider two contextual sentences before and after a potential claim/argument, which enabled them to correctly classify sentences that would have otherwise been left out. For instance, in a sequence of sentences "The flow of arms to Ukraine is enormous. This is unacceptable," the second sentence can only be coded as a claim against arms deliveries if contextual information is accounted for. Contextual information was also used later to train a classification model. Here, our communication science application scenario deviates from the standard argument mining task in NLP which usually operates on decontextualized sentences only (Jurkschat et al., 2022). This inclusion of context during manual coding needs to be reflected in modifications to procedures of machine input to the argument mining process (cf. Section 3.4).

Lastly, the codebook specified how to label those sentences that contained one actor contradicting another actor's position on arms deliveries, e.g.: "Plötner complained that the media were becoming more focused on tank deliveries instead of focusing on future relations with Moscow." The codebook referred to such sentences as *onion-structured* claims/arguments and instructed coders to always label the (alleged) stance of an actor on *arms deliveries* (here, media) rather than the stance of an actor on *another actor's position (*here, Plötner). To enable controlling for the stance of the onion-structured claims/arguments at the further stages, such sentences were labeled with a separate code (for the full codebook, see Appendix 2 of the Supplementary Information file in the OSF). We hypothesize that logical contradiction and critique of well-formed arguments might be difficult to grasp for the machine from a few samples only. In our experiments, we, thus, investigate whether the original labels (commenters' stance) or the labels with a swapped stance (actor's position commented on), while keeping the argument concept as labeled, are better suited for automated processing using FSL.

**3.3 Coding process**

The coding for arguments and claims was carried out by four coders in parallel. For comparable coding units, the texts were split into sentences using the NLTK tokenizer (Bird et al., 2009). After attending a training session where the coders familiarized themselves with the codebook and practiced assigning codes to data, they performed three rounds of coding on three sets of articles randomly sampled from the final corpus. Each coding round was followed by a group discussion during which inconsistently coded sentences were analyzed and coding rules clarified.

Table 3. Results of the inter-coder reliability tests for each of the coding rounds

|  | n coders | n articles | n total sentences | Kalpha relevant sentences | n relevant sentences | Kalpha claim/argument |
|---|---|---|---|---|---|---|
| Round I | 4 | 50 | 5242 | 0.594 | 119 | 0.497 |
| Round II | 4 | 70 | 6729 | 0.634 | 81 | 0.667 |
| Round III | 4 | 70 | 5073 | 0.814 | 246 | 0.662 |

Table 3 reports the results of the inter-coder reliability tests for each of the coding rounds. Due to the complexity of the coding task, inter-coder reliability was assessed in two steps. In the first step, coders needed to agree on whether each specific sentence from a dataset contained a stance regarding weapons deliveries to Ukraine and hence was relevant for further classification (for the agreement scores, see Table 3). In the second step, the relevant sentences on which the majority of coders agreed in each round were classified as either argument for/against or claim for/against weapons deliveries. As can be seen from Table 3, in the first round of coding, the coders achieved "moderate" (Hughes, 2021, p. 417) agreement scores while in the last two rounds, the scores improved to >0.6, which is generally regarded as "significant" (Cabrio & Villata, 2018, p. 5428) or "substantial" (Hughes, 2021, p. 417) agreement.

When using the labeled data to train a model, the majority rule was applied again to address the issue of imperfect inter-coder agreement. Particularly, only sentences on which most coders agreed at a second coding step qualified for the training dataset (this meant the agreement of two or three coders depending on how many coders labeled a sentence as *relevant* in the first step). This approach guarantees high-quality data for training purposes but also allows for some degree of disagreement among the coders so that not only *easy* cases of claims and arguments are picked to train and later assess the classifier.

Table 4. Number of observations per label and data set in dependence of stance-swapping for argumentative sentences labeled as onion-structured

| Label | Original | | | Onion | | |
|---|---|---|---|---|---|---|
| Set | train | dev | test | train | dev | test |
| **argumentfor** | 46 | 4 | 13 | 49 | 4 | 14 |
| **argumentagainst** | 45 | 7 | 14 | 42 | 7 | 13 |
| **claimfor** | 101 | 13 | 24 | 101 | 13 | 23 |
| **claimagainst** | 81 | 14 | 19 | 81 | 14 | 20 |
| **nostance** | 1091 | 155 | 317 | 1091 | 155 | 317 |

For later modeling, it is beneficial if the imbalance of stance to nostance sentences is not too large. Therefore, we randomly sampled a set of nostance sentences from the total set so that they make up 80 percent of the total dataset. For the quality assessment of the experiments in Section 4, we divide the dataset into three parts: a train (70%), a dev (10%), and a test (20%) set. Table 4 provides the final number of sentences for each category depending on whether the labels are onion-swapped.

### 3.4 Language models

We compare different approaches to fine-tuning PLMs for text sequence classification in automated content analysis. The four underlying approaches are explained in Table 5 regarding their conceptual idea and in the following with respect to their methodology. Please see Appendix 3 of the Supplementary Information file in the OSF for detailed information and explanation regarding the concrete implementation of the models.

All our experiments are based on the XLM-RoBERTa model in the large variant from Huggingface's transformer package (Wolf et al., 2020), which has a file size of 2.1 GB. A decisive advantage of the model is its multilingualism so that all presented analyses are easily

Table 5. Comparison of different approaches for fine-tuning PLMs.

| Approach | FSL | PEFT | Description |
| --- | --- | --- | --- |
| Full fine-tuning (FT) | | | Updating all parameters of the pre-trained language model (PLM) by learning relations between features and codes from a coded dataset. |
| Near-domain fine-tuning | X | | Updating all parameters of the PLM by learning relations between features and codes from a coded dataset that has similarities to the dataset of interest in at least one component. Near-domain fine-tuning is performed as a preliminary step of actual fine-tuning on the dataset of interest. |
| Pattern-exploiting training (PET) | X | | Updating all parameters of the PLM by learning relations between features and semantic representations of codes from a coded dataset using a language modeling objective instead of the standard classification objective. |
| Adapters | | X | Freezing all parameters of the PLM, adding and updating only new parameters by learning relations between features and codes from a coded dataset. |

*Note.* FSL = few-shot learning, PEFT = parameter-efficient fine-tuning.

transferable to other languages. To account for the context of sentences during coding, we construct the input to our standard models (without PET) as follows

$$\text{Input} := [\text{target sentence}] \text{</s>} [\text{context before}] \text{</s>} [\text{context after}]$$

where </s> represents a special separator token. The model has a maximum input sequence length of 512 tokens. Longer inputs are right-truncated to 512 tokens, so that the context after would be cut first, then context before, then the target sentence. The order ensures preservation of the most informative parts for overly long inputs. On our dataset, shortening affected only a few nostance target sentences. Placing the target sentence at the beginning also facilitates the model to particularly focus on the first tokens. Positioning between contexts, in contrast, would result in greater variance in the positioning of the tokens most relevant to the task.

As a baseline model, we use standard fine-tuning with a learning rate of 5e-6 together with a linear learning rate scheduler with warm-up. This setup makes fine-tuning less prone to overfitting. By empirical testing, we decided to run the training for 30 epochs.[3] As a direct comparison, we consider fine-tuning with identical parameters, using an already fine-tuned model on a near-domain dataset. For this, we use the well-known UKP-SAM dataset containing argumentative English-language sentences from eight different topics together with a pro/contra/neutral stance label (Stab et al., 2018; Reimers et al., 2019) and train the XLM-RoBERTa model with a learning rate of 5e-6 for two epochs.

*3.4.1 Pattern-exploiting training (PET)*

Since fine-tuning requires a certain amount of data and a common problem in practical applications is the small amount of labeled data, we investigate to what extent the state-of-the-art FSL model PET (Schick & Schütze, 2021) adds value to the correct identification of argumentative sentences. PET requires an additional manual step for its application, namely the definition of pattern-verbalizer pairs (PVPs). We use two types of PVPs, one somewhat naive and one more elaborate, and train the model for 10 epochs using a learning rate of 1e-5.

*3.4.2 Parameter-efficient fine-tuning with adapters*

As a resource-efficient alternative to full fine-tuning, we also investigate the use of adapters in our experiments. There are many possibilities to use different adapter architectures. The most frequently investigated architecture is the *pfeiffer* adapter (Pfeiffer et al., 2020), which produces consistently promising results. So, we will concentrate on this architecture for our analysis. We train it with a reduction factor of 16 and a learning rate of 5e-5 over 30 epochs.[4] These values can be seen as a reasonable default that has reliably provided solid to very good results in previous studies. Hence, we use them for all bottleneck adapters presented in this paper.

---

[3] We tested the standard fine-tuning comparing the more common setup of 10 epochs vs. 30 epochs. Training for 10 epochs leads to significantly worse results (see Appendix 6 of the Supplementary Information file in the OSF).
[4] Adapters are even less prone to overfitting than fully fine-tuned models for higher learning rates and higher numbers of epochs due to the freezing of the initial model.

*3.4.3 Combining parameter-efficient fine-tuning and pattern exploiting-training*

As a combination of the ideas of adapters and PET, we implement a new adapter with a PET-like classification head. For this, we only investigate the use of the naive PVP and combine a standard bottleneck adapter with our implementation of the PET-like classification head. Due to the modularity of adapters, it is also possible to combine this implementation with near-domain standard/adapter fine-tuning. For reasons of complexity, we only consider the combination of near-domain adapter fine-tuning and the PET-like head in our experiments.

**4 Identifying argumentative sentences in the news media debate**

We conducted three major experiments that analyze different facets and properties of model decisions to determine an optimal workflow for FSL, which are described in detail in the following three subsections.

In the first experiment, we investigated the influence of two preprocessing steps on the validity of the results. We compare the standard fine-tuning as a baseline with additional fine-tuning on the near-domain fine-tuned model. To shed light on how supervised learning can grasp complex argumentative structures, we compare the performance of both models with respect to whether sentences labeled with onion-structured contexts can be predicted more accurately with their actual or their swapped stances (for/against) (cf. Section 2.1). During our initial experiments, we observed systematic misclassifications related to the mention of certain person names. Thus, we also investigate this potential bias by randomly replacing all person names with other names before training. We compare the total of eight model variants in terms of overall macro-F1, precision, and recall, as well as the corresponding class-specific validity measures. For all models, we perform five repetitions each, so that we can also assess the reliability of the validity values via the models' uncertainty. The second experiment examines the application of all models and methods presented in Section 3.2 in a few-shot setting by evaluating different models at varying training set sizes. In addition to assessing the validity of the predictions, we again estimate their reliability in relation to specific models or numbers of training data by repeating them five times. In the third experiment, we test the most promising model from the second experiment with respect to its suitability in the real few-shot scenario.

These experiments provide insights into the validity, reliability, and replicability of the results of various state-of-the-art NLP models (including our own newly introduced combination of adapters and PET), as well as two preprocessing decisions strongly related to their application on the task of argument mining. In addition, in Section 4.4, we discuss the feasibility and model sizes in combination with the replicability of the applied models and their practical applicability in the field of communication science.

## 4.1 Biasing models away from person's stance

Initially, it may sound fallacious to try to prevent a model from learning that certain individuals are closely associated with a certain position. Indeed, for human interpretation the person with whom a statement appears matters, as well (Westerwick et al., 2017, p. 346). However, these individual positions may distort the model too much, so in the example of Annalena Baerbock (German Federal Minister of Foreign Affairs) — who is 90 percent of the cases connected to a positive stance on weapons delivery in our dataset — the person may predominate over the remaining elements of the sentence and only positive stance is predicted for those sentences. This poses several dangers with respect to the generalizability of the resulting model. If a person's position changes over time, but our model was only trained on data up to the time of the change, we might consistently misclassify statements from that person. Similarly, if our dataset only contains sentences with Annalena Baerbock as the actor, but the sentences to be predicted then also contain sentences with her as a reference, then the model will already be biased with respect to the classification of the sentence purely by the occurrence of this person.

For this reason, in the following, we consider the effect of a regularization preprocessing step that is intended to remove the information on the association of single individuals with positions. To do this, we use named entity recognition (NER) using the model *ner-german-large* from the flairNLP package (Akbik et al., 2019) to identify individuals in our dataset and replace each occurrence of an individual with a random entry from the list of all occurring person entities. We also make sure that repeated occurrence of the same entity in one single sentence is replaced with the same (random) entity. In the following experiment, we distinguish between models with respect to the preprocessing step *Person* with the characteristics *original* and *shuffled*, where *original* means no changes of the initial data and *shuffled* refers to the application of the described procedure.

Another preprocessing step we consider is the stance reversal of sentences labeled as onion structured. The idea here is to investigate to what extent PLMs can recognize stance-reversing phrases (e.g., negations, distancing). In total, the data contains 49 sentences labeled as onion-structured (cf., Section 3.1). In the experiment, we distinguish between the original dataset (original) and the dataset with swapped labels (onion).

We consider the two presented models with full fine-tuning, i.e., standard fine-tuning (FT) and near-domain fine-tuning as a pre-step of full fine-tuning (FT SAM) as described in Section 3.2. Instead of a fixed number of epochs, we follow a common approach by training for 50 epochs on our labeled dataset and selecting the best epoch with respect to the macro-F1 score using the dev set. We use this best model for evaluation on the test set.

Table 6 shows the results of the eight combinations from the investigated models and the two preprocessing steps. We examine macro-F1 score and accuracy for general performance assessment, while additionally reporting class-dependent F1 score, precision, and recall.

This shows that the model with the best overall performance is FT SAM without onion-swap but with person shuffling with a macro-F1 score of 66.0 percent and an accuracy of 91.6 percent. It should also be mentioned that accuracy is a somewhat problematic metric here, since due to the imbalance of labels the majority baseline to always predict nostance would already achieve an accuracy of 80.4 percent.

It turns out that the individual classes (categories) differ greatly in their performance. The label nostance is classified best (best F1: 0.997), while argumentfor performs worst (best F1: 0.395). This observation can be well described by the number of training data per label. Arguments are consistently predicted with higher precision than recall. For claims, the pattern is the other way around, with one exception. In an error analysis of a single sample run of the best-performing model, we found that this can be explained by the presence or absence of argument markers: while their presence leads to the high precision of arguments, their absence causes the models to assume a claim. Thus, the most frequent error in this sample run is that a claim is predicted where an argument is present (10/26 errors), while the stance is still correctly predicted. Moreover, among correctly classified arguments, 7 out of 11 have either argument markers or trigger words such as *argument* or *consensus* (for the comparison of correctly and incorrectly classified arguments, see Appendix 4 of the Supplementary Information file).

Table 6. Comparison of performance measures for full fine-tuned models based on two preprocessing decisions.

| Person | original | | | | shuffled | | | |
|---|---|---|---|---|---|---|---|---|
| Label | original | | onion | | original | | onion | |
| Model | FT | FT SAM | FT | FT SAM | FT | FT SAM | FT | FT SAM |
| **Overall:** Macro-F1 | 0.594 ± .027 | 0.634 ± .036 | 0.608 ± .067 | 0.627 ± .039 | 0.585 ± .040 | **0.660 ± .027** | 0.566 ± .048 | 0.635 ± .026 |
| Accuracy | 0.906 ± .005 | 0.908 ± .008 | 0.902 ± .009 | 0.908 ± .009 | 0.901 ± .007 | **0.916 ± .011** | 0.893 ± .010 | 0.906 ± .011 |
| **argumentfor:** F1 | 0.243 ± .077 | 0.339 ± .111 | 0.322 ± .170 | 0.356 ± .061 | 0.271 ± .138 | **0.395 ± .085** | 0.198 ± .166 | 0.391 ± .036 |
| Precision | 0.317 ± .020 | 0.366 ± .112 | 0.360 ± .227 | 0.412 ± .087 | 0.311 ± .159 | 0.460 ± .114 | 0.215 ± .176 | **0.465 ± .084** |
| Recall | 0.215 ± .102 | 0.323 ± .123 | 0.314 ± .173 | 0.329 ± .073 | 0.246 ± .132 | **0.369 ± .132** | 0.186 ± .160 | 0.343 ± .029 |
| **argumentagainst:** F1 | 0.355 ± .063 | 0.478 ± .061 | 0.467 ± .087 | 0.479 ± .098 | 0.386 ± .064 | **0.507 ± .069** | 0.483 ± .147 | 0.501 ± .032 |
| Precision | 0.456 ± .097 | 0.576 ± .108 | 0.549 ± .065 | 0.613 ± .058 | 0.654 ± .195 | **0.666 ± .094** | 0.537 ± .188 | 0.629 ± .122 |
| Recall | 0.293 ± .053 | 0.413 ± .050 | **0.443 ± .153** | 0.400 ± .107 | 0.280 ± .050 | 0.413 ± .065 | **0.443 ± .123** | 0.429 ± .045 |
| **claimfor:** F1 | **0.689 ± .023** | 0.649 ± .023 | 0.573 ± .057 | 0.588 ± .041 | 0.597 ± .060 | 0.670 ± .031 | 0.531 ± .020 | 0.604 ± .052 |
| Precision | **0.632 ± .021** | 0.574 ± .018 | 0.578 ± .080 | 0.537 ± .044 | 0.527 ± .083 | 0.607 ± .047 | 0.462 ± .017 | 0.570 ± .097 |
| Recall | **0.758 ± .031** | 0.750 ± .053 | 0.574 ± .051 | 0.652 ± .048 | 0.700 ± .049 | 0.750 ± .026 | 0.626 ± .044 | 0.661 ± .058 |
| **claimagainst:** F1 | 0.699 ± .024 | 0.719 ± .035 | 0.695 ± .059 | 0.726 ± .021 | 0.685 ± .024 | **0.744 ± .038** | 0.637 ± .039 | 0.694 ± .045 |
| Precision | 0.602 ± .035 | 0.642 ± .029 | 0.603 ± .076 | 0.648 ± .036 | 0.584 ± .018 | **0.676 ± .043** | 0.607 ± .062 | 0.616 ± .035 |
| Recall | **0.838 ± .049** | 0.819 ± .047 | 0.836 ± .079 | 0.827 ± .034 | 0.829 ± .038 | 0.829 ± .038 | 0.673 ± .034 | 0.800 ± .084 |
| **nostance:** F1 | 0.991 ± .005 | 0.997 ± .002 | 0.990 ± .003 | 0.992 ± .003 | 0.992 ± .004 | 0.990 ± .004 | 0.989 ± .005 | 0.989 ± .005 |
| Precision | 0.979 ± .006 | 0.973 ± .006 | 0.976 ± .004 | 0.980 ± .005 | 0.977 ± .004 | 0.981 ± .012 | 0.979 ± .006 | 0.977 ± .010 |
| Recall | 0.985 ± .001 | 0.985 ± .003 | 0.983 ± .001 | 0.986 ± .001 | 0.984 ± .002 | 0.985 ± .005 | 0.984 ± .002 | 0.983 ± .003 |

*Note.* Mean values ± standard deviation.

Overall, the results from the first experiment do not show clear findings regarding the question of whether it is better to use the original coded labels or the onion-swapped labels for the models. In contrast, person shuffling leads to better results, since the model seems to focus more on the actual argumentative part of a sentence than just the person stating it. Thus, for all further analyses, we will use the person shuffle preprocessing. In addition, the results show that pre-training with a near-domain dataset improves the performance of the models.

**4.2 Model comparison for different training set sizes**

Since in real applications, it is often not possible to label that much data (quickly) and since it is interesting to know what proportion of training data leads to sufficiently good predictions, we investigate eight different models with respect to their FSL performance regarding the overall macro-F1 score. For this purpose, we sample fixed train sets with proportions of 2.5, 5, 10, 20, 30, 50, 70, and 100 percent of the complete train set stratified by the labels (for absolute numbers of labels per train sample see Tables A2 and A3 in Appendix 5 of the Supplementary Information file in the OSF). In our FSL setting, we do not consider a dev set, but train with previously defined hyper-parameters as shown in Section 3.2, since such a dev set would not be available in a real FSL setting as well. We use person shuffle preprocessing and compare results from the dataset with the original labels.[5] All models are applied for five repetitions to assess the reliability of the models in the few-shot setting.

In order to improve readability, we compare the four models based on general approaches in Figure 1 on the left and the four specialized FSL approaches on the right. In addition, from the

---

[5] We also ran the experiment for the onion-swapped labels. This yield comparable results, but no consistent improvement or decrease in performance.

Figure 1. FSL performance evaluation for different models and target labels

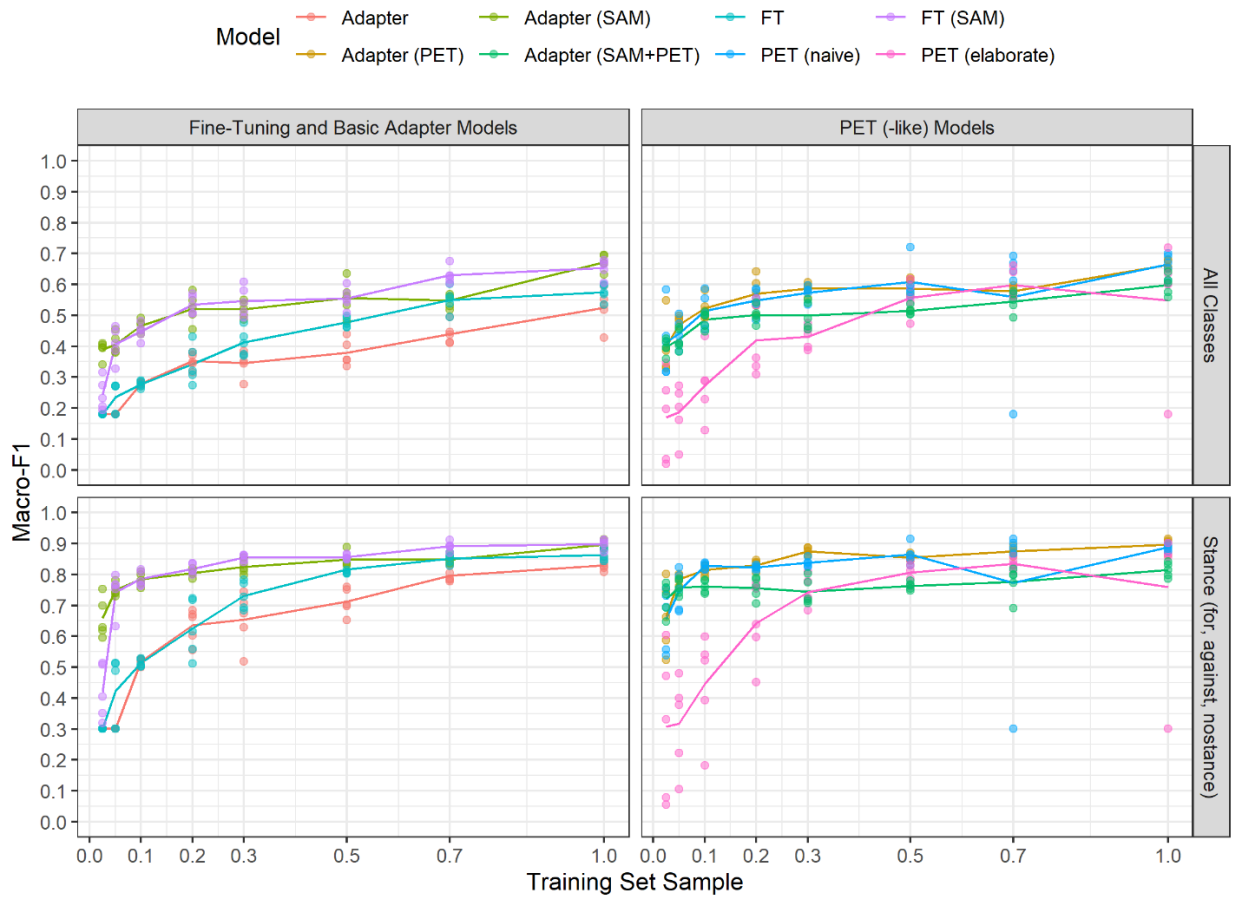

*Note.* Dots represent individual runs; curves represent mean values.

original predictions on the 5-class problem in the top row, we also computed the performance for a pure stance prediction (for, against, nostance) in the bottom row.

Confirming the result from the first experiment, the use of near-domain pre-training improves the performance of the models. In the FSL setting, we can see that for the adapter-based approaches, there is nearly a constant difference of 20 percentage points between the macro-F1 score for basic adapters and the additional use of a pre-trained adapter. The model *FT (SAM)* and *Adapter (SAM)* perform best among the general approaches. While the full fine-tuning on average provides slightly better results for the FSL settings with 30, 50, or 70 percent, the adapter-based approach achieves a macro-F1 score above 67percent on the complete data set for both label types, which is better than the best model from Section 4.1, where also the dev set is used. The full fine-tuning after near-domain pre-training achieves similar scores as in the first experiment with a macro-F1 score of 0.653 (original) and 0.652 (onion). Considering the reliability of the validity estimation, both models can be considered equally well-suited for the task.

In the comparison of the standard and the pre-trained fine-tuning, there is not such a *constant* difference as for the adapter-based approaches, but the performances tend to become closer as the training data grows. Therefore, the pre-training with the SAM dataset is particularly beneficial in the FSL setting. Although we know that adapters perform on par on full data for several tasks (cf., Section 3.2), here the full fine-tuning approach performs slightly better than the basic adapter for growing training data.

PET using the naive PVP, and our newly proposed PET head adapter perform overall on par, while on the full training data, our PET head performs best. The more elaborate PVP for a few training data is shown to perform poorly and at the same time to result in a high uncertainty in the validity of the results. Two outliers stand out for the naive PET for 70 percent and for the more elaborate PET for 100 percent of the training data. In both cases, the models end up predicting only nostance. For very few data, the combined *Adapter (SAM+PET)* model works quite well, but for a growing database, the performance of the model is almost consistently 10 percentage points below the *Adapter (PET)* model without SAM pre-training. Among the best-performing models (PET (naive), Adapter (PET), Adapter (SAM), and FT (SAM)) our newly proposed PET head has the lowest uncertainty in validity, i.e., the highest reliability of the results. Even for just five percent of the train set, our PET head achieves a macro-F1 score of almost 50 percent in the onion-swapped case. Finally, it can be seen that a combination of the SAM adapter and a PET head does not achieve the desired enhancing effect, but leads to a worsening of the results — (probably) due to the different training targets of the SAM adapter (classification) and the PET head (language modeling).

As already shown by the error analysis in the first experiment, the difficulty of the task consists in particular in the prediction of claim vs. argument, while the stance is estimated very reliably (macro-F1 of 90 percent on the complete data set). It should be noted that these models were not explicitly trained on the task of 3-class classification, and explicit training on this task could lead to (even) better results.

**4.3 Argument mining in the true few-shot scenario**

In the previous section, our adapter-based model with PET head proved to be the most promising in the FSL setting. In the following, we compare this model with respect to its validity and replicability in the true few-shot (TFS) setting. For this, we sample the train set not only once and stratified but ten times without the condition of stratification. By this, we simulate the process of coding random sentences without any prior knowledge of the actual label distribution. We examine the same sample proportions as in Section 4.2.

Figure 2 shows how F1 score, precision, and recall relate to the amount of training data. The results show a difference in the overall macro-F1 score over ten repetitions of more than 30 percentage points between the worst and the best repetition in the case of only 2.5 percent of the training data. This uncertainty reduces to less than 20 percentage points for the FSL settings using 10, 20, 30, or 50 percent of the data, while for 70 percent it shrinks to nearly ten percentage points. As expected, the uncertainty decreases for higher proportions of the training set. It should also be noted that we cannot eliminate model uncertainty in this experiment.

Figure 2. TFS performance evaluation of the PET head for differing subsets of the train set

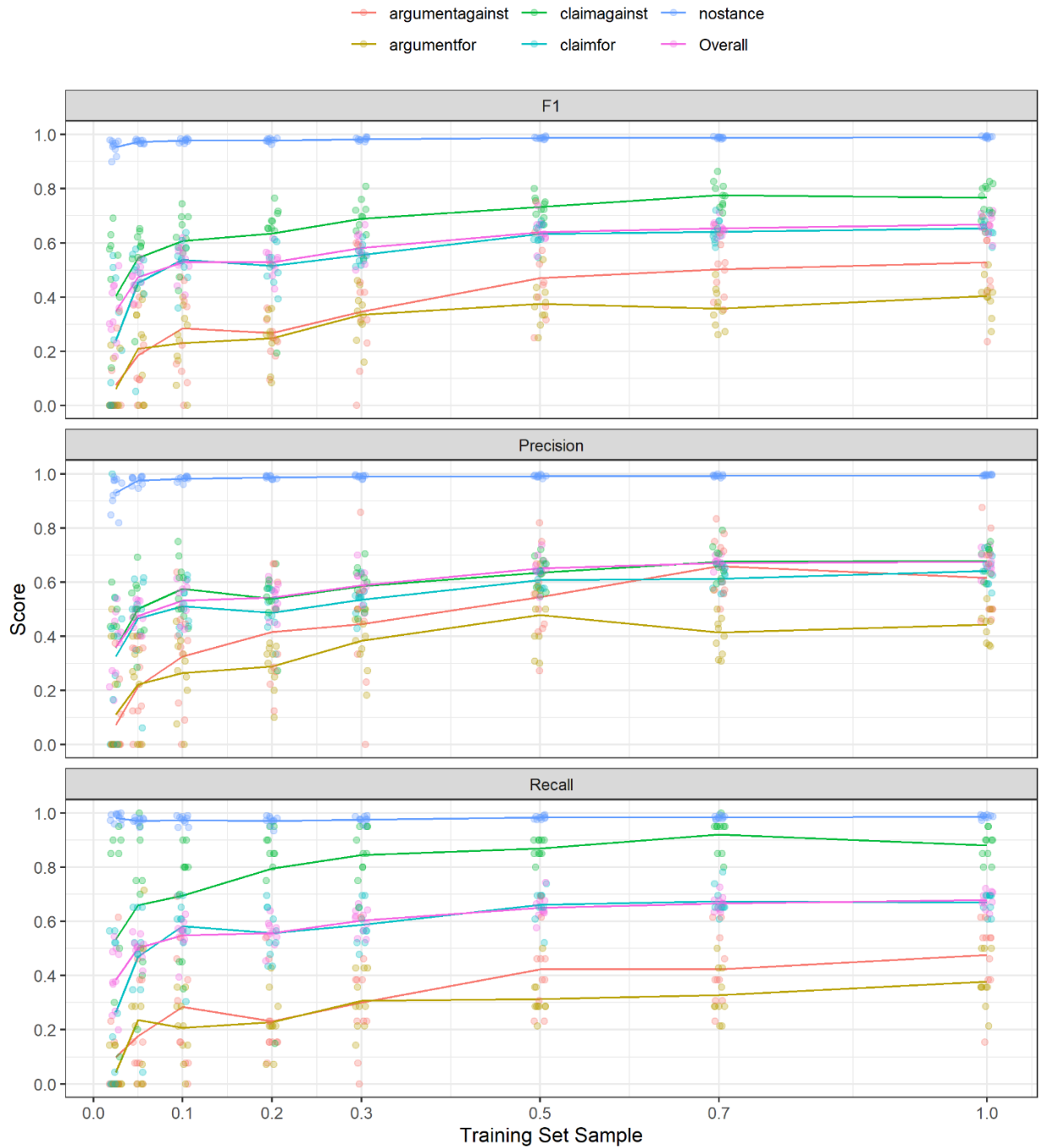

*Note.* Dots represent individual runs; curves represent mean values. The dots are slightly jittered in their horizontal dimension for better readability.

In Section 4.2, we found that even using the identical subsets for five repetitions in the case of 50 percent of the training data results in a range of five percentage points between the best and worst macro-F1 score for the PET head model we investigate here. This suggests that for a sufficiently large data set, the greater uncertainty comes from the modeling rather than from the data set sampling.

In addition, we can confirm the findings from Section 4.1 that, in principle, for arguments, the precision is higher than the corresponding recall, while for claims it is the other way around. The replicability of the results using the same method, but different samples of the same data basis is quite limited for this difficult task in the TFS setting. This is illustrated by the example of the reliability of the different validity measures, F1, precision, and recall, of which especially the latter two partly show large uncertainties.

**4.4 Feasibility and reproducibility of the trained models**

Besides validity (cf., macro-F1 scores in Sections 4.1, 4.2, and 4.3), reliability (cf., repetitions of models in Sections 4.1 and 4.2) and replicability (cf., different train sets in Section 4.3), feasibility and reproducibility are important properties for the (meaningful) applicability of automated models – not only for content analysis.

In Table 7 the file sizes of the models are given, whereby sharing them guarantees the complete reproducibility of the results. Here it is assumed that the PLM XLM-RoBERTa is already available. All adapter approaches require only 12 MB each for storing all additional learned parameters of the model plus 4 MB for the prediction heads, whereas the parameters of the initial model remain the same. This makes them easy to share and just as easy to reproduce their predictions, especially in studies with numerous models. Combined with the finding that near-domain pre-training has a positive impact on the validity of the results, makes the (re)use of such near-domain adapters attractive.

Table 7. Sizes and training time

| Near-domain pre-training: | Size | Runtime (100 %) | Runtime per epoch (100 %) | Runtime (2.5 %) |
|---|---|---|---|---|
| SAM | 2.1 GB | 54 min | 26.35 min | - |
| SAM Adapter | 12 MB (+ 4 MB) | 455 min | 15.17 min | - |
| **Fine-tuning:** | | | | |
| FT | 2.1 GB | 52 min | 1.73 min | 77 sec |
| Adapter | 12 MB (+ 4 MB) | 37 min | 1.23 min | 53 sec |
| Adapter (PET) | 12 MB (+ 4 MB) | 45 min | 1.50 min | 65 sec |
| PET | 2.1 GB | 54 min (+ 8 min) | 5.40 min | 77 sec (+ 8 min) |

*Note.* The pre-training is done on the UKP-SAM dataset (cf., Section 3.2) while fine-tuning is performed using our presented dataset (cf., Section 3.1). The file sizes in parentheses refer to the classification head; the times in parentheses reflect the time PET needed to perform the language model objective in addition to the classification training. The experiments were conducted on a local machine using an NVIDIA GeForce RTX 3090 with 24GB.

In addition to the file size, the runtimes for training the models in the FSL scenarios are given for 2.5 and 100 percent of the train set. The near-domain pre-training requires considerable computation time due to the application to the rather large UKP-SAM dataset. However, we assume here that in practice one of these near-domain datasets and a pre-trained model are already available so the runtime in this case is not important for the actual application. The computation time for fine-tuning of our dataset, in contrast, is relevant and varies for the different models. At 5.40 minutes per epoch, PET is the most expensive, while the adapter-based approaches are significantly faster to compute at 1.23 minutes and 1.50 minutes per epoch, respectively. Thus, adapters form a viable option for application in low-resource environments.

Compared to the adapters with standard classification heads, the PET head, at 45 minutes on the entire dataset, requires slightly more training time. However, its reliable and performant application without the need for tuning offers the possibility to obtain a decent model without running the training multiple times with different parameters.

## 5 Discussion

We investigate the use of PLMs for the identification of claims and arguments in (semi-) automated content analysis of media debates with little training data for a typical communication science research scenario. The application of three different experiments allows us to assess the validity of the results depending on pre-processing steps, model choice, and amount of training data. In addition, our experiments provide insights into the reliability, replicability, and reproducibility of the investigated model architectures, and allow us to draw some conclusions on the usefulness of this method for communication research.

### 5.1 General findings

In general, we found that all models studied are better at identifying manifest arguments and worse with latent argumentation, at least with the present amount of training data. The swapping of data labeled as onion-structured does not lead to a consistent difference in validity. A promising preprocessing step for the analysis of data in combination with stance is a random shuffling of individuals, which deprives the models of the information of the individuals' stances but puts more weight on other elements of the sentences. As a result, the validity of the results improved. Furthermore, it is shown that near-domain pre-training leads to better results than training on just in-domain data.

Despite their impressive performance, the models could not achieve satisfying values for all categories (aka classes) from a communication research (or machine learning) perspective. And yet, the validity — as well as their reliability over repetitions — should be sufficient to allow meaningful analyses of trends and distributions in the material based on the automatic classification alone.[6] In the category with the worst performance, arguments for weapons deliveries, most errors were a result of the models mistaking arguments for claims, with the stance almost always correctly predicted. In other words, even at this stage of training, the method explored here could already be used to rebuild all those previous studies which manually coded the distribution of pro- and contra-stances within media texts at a fraction of the cost. And even if at this point, manual coding may still be more reliable in identifying arguments for weapons deliveries, the time (and cognitive effort) required for this would be reduced substantially by first extracting the relevant sentences from the sample automatically using the proposed models.

**5.2 Model choice**

In the case of near-domain pre-training, adapters perform on par with full fine-tuning, while without pre-training, full fine-tuning becomes increasingly better than the selected basic adapter as the number of training data increases. Our newly proposed PET head performs on par with the original PET model in terms of validity but produces more reliable predictions. We were able to show that PET has a high uncertainty in the results when performed repeatedly on the same data, especially in the FSL setting and for too elaborated PVPs. Overall, for the presented task, our PET head is the best model choice based on validity, reliability, and reproducibility (parameter efficiency). For an application in low-resource environments, we recommend the use of adapters in combination with pre-training, for which also the manual step of engineering pattern-verbalizer pairs can be omitted.

---

[6] Based on extensive experiments, Wiedemann (2019) concludes, "already moderate individual classifier performance regarding common evaluation measures such as F1 or Cohen's kappa provide sufficiently accurate results to validly predict proportions and trends in a collection" (p. 155).

## 5.3 Limitations

In our experiments, we have chosen selected and established default hyper-parameters. For adapters, we restricted our analysis to the best-known standard variant, the *pfeiffer* adapter in combination with a compression rate of 16. We did not investigate the robustness of the models with respect to the choice of parameters beyond the number of epochs in the FSL setting. We tested 10 epochs (a commonly chosen non-tuned parameter for full fine-tuning in NLP literature) against 30 epochs. Training with fewer epochs showed strong drops in the validity of the results (cf., Figure A1 in Appendix 6 of the Supplementary Information file in the OSF).

During the coding process, it became apparent how difficult the task is — even for human coders — so the question of possible error propagation compared to actual true labels (as a kind of *platinum* standard) arises. We have tried to counteract this by removing examples that cannot be clearly labeled to ensure the quality of the training data. However, it is unclear to what extent this decision might lead to an overestimation of the validity of the results due to a potential simplification of the task.

## 5.4 Outlook

In further experiments, we will investigate the robustness of the PET head for different parameters and base architectures on different datasets. We aim to improve the implementation in such a way that PVPs might no longer be necessary (cf., von Werra et al., 2022; Mahabadi et al., 2022). Further potential for improvement may lie in the choice of the loss function used.

Possibly the use of a triplet loss (cf., Sosnowski et al., 2022) in combination with certain data augmentation strategies could be promising.

Concerning the coding process, we plan to conduct several investigations. In terms of Active Learning (cf., Wiedemann 2019, Markus et al., 2023), it is interesting to find out whether there is potential for improvement during the coding process to draw further coding examples that are particularly useful for the model, using a more elaborate strategy than simple random sampling. We also aim to expand the coding of arguments beyond the sentence level by identifying premises among contextual sentences. This will not only allow us to provide a more complete picture of the German media debate concerning weapons deliveries to Ukraine but, more importantly, the inclusion of premises may help improve the models' performance in identifying arguments.

Moreover, since near-domain pre-training has proven to be useful and it has been shown in another work that stance prediction is not topic independent (Reuver et al., 2021), it would be desirable to find the best fitting near-domain dataset, which is already thoroughly labeled, based on automated criteria and similarities to a weak-labeled dataset of interest. For this, we plan to develop a kind of similarity metric using different near-domain datasets that strongly correlate with their usefulness in being used as a near-domain dataset, so that it would be appropriate for finding a suitable dataset for pre-training.

Overall, by conducting these three experiments using the latest developments in NLP on a concrete communication science research case, we were able to address the three critical research gaps identified by Baden et al. (2022). Due to the close collaboration between computer and communication scientists within this project, it was possible to keep a very close eye on the validity of the classified constructs both by conducting manual error analyses but also by testing the impact of different preprocessing steps such as the de-biasing of the data by randomizing names. Furthermore, with the advent of multi-language transformer models such as the XLM-RoBERTa used here, we are no longer limited to the analysis of English texts (though of course languages without a sufficient amount of digital text available are still out of reach). And finally, we could show that by coding claims and arguments on a sentence level using the surrounding sentences as a context both for the human coders and the models, these methods can now be employed successfully for more complex constructs, greatly increasing the range of possible communication research questions that might be addressed.

## Acknowledgements

This work was funded by the Deutsche Forschungsgemeinschaft (DFG) as part of the project *FAME: A framework for argument mining and evaluation* (project no. 406289255) and by the Bundesministerium für Bildung und Forschung (BMBF) as part of the project *FLACA: Few-shot learning for automated content analysis in communication science* (project no. 16DKWN064B).

# Supplementary Information File

## Appendix 1. Media outlets constituting the final corpus

"Bild am Sonntag"
"Bild Bund"
"Bild.de/plus"
"der Freitag"
"DER SPIEGEL"
"Der Tagesspiegel"
"die Tageszeitung"
"Die WELT"
"Die Zeit"
"FAZ Plus"
"FAZ.net"
"Focus"
"Frankfurter Allgemeine Zeitung"
"Frankfurter Allgemeine Sonntagszeitung"
"Frankfurter Rundschau"
"Handelsblatt"
"Neue Zürcher Zeitung"
"Neue Zürcher Zeitung am Sonntag"
"Spiegel Online"
"Spiegel Plus"
"Stern"
"stern plus"
"Süddeutsche Zeitung"
"süddeutsche.de"
"Welt am Sonntag"
"Welt.de"
"WELTplus"
"zeit online"

# Appendix 2. Codebook[7]

Coding task #1

*Thematic scope of the coding task*

**###** In the frame of this coding task, we are interested in published viewpoints regarding **the (in)appropriateness of arms deliveries to Ukraine**. This includes any type of arms, e.g., defensive/heavy arms; any type of delivery schemes, e.g., direct/in cooperation with other countries; and any other aspects not mentioned here, e.g., purchase from the manufacturer/donation from the Bundeswehr [The Armed Forces of the Federal Republic of Germany] stocks, etc.

**###** As we work with the German newspaper discourse whose target audience are German readers, we consider general mentions of arms deliveries to Ukraine as relevant. Non-relevant are only those mentions of arms deliveries that name specific non-German actors, e.g., the US, France, Poland (i.e., arms deliveries that are explicitly NOT from Germany). We also consider relevant the mentions of arms deliveries from the entities and organizations that include Germany, e.g., the EU, NATO, the West, etc.

**###** Anything that is **NOT** arms deliveries should **NOT** be coded, e.g. providing military training, sending helmets/humanitarian aid/troops, "closing the sky," the so-called "Ringtausch" [ring exchange], etc.

*Units of coding*

**###** To code we use Word's commenting function which can be accessed through the "Review" tab. The unit of coding is a sentence in its entirety without the last punctuation mark, i.e., during the coding, the commented text should never include the last punctuation mark, e.g.:

We should send weapons to Ukraine.

He said: "We should send weapons to Ukraine."

How could you even think of sending weapons to Ukraine!

"We must be prepared to deliver more weapons in the future", said Strack-Zimmermann.

If a sentence to be coded is the first one, we leave a comment in the following way:

text: Arms should be delivered to Ukraine quickly.

All comments must be added in this way so that the coding done by different people can be compared automatically. It is also important not to change anything in the document where coding takes place.

**###** Each article to be coded starts with a unique id. This identifier should be commented with the word "id." For example:

id: ASV-HABO20220509-EXTASV-HABO-235292223 (CODE: "id")

---

[7] The codebook is presented in its original English version. At the same time, it includes some of the claim/argument examples from the German language data used in the study. Those example sentences were translated from German to English for the sake of consistency.

*Types of codes*

In the frame of this coding task, we assign the following codes at the sentence level:

111 – argument for

112 – argument against

121 – claim for

122 – claim against

888 – onion structure

*Arguments*

**###** Argument is a prescription/demand/viewpoint regarding (in)appropriateness of arms supplies from Germany to Ukraine **with explanation**(s) of such position.

  Argument for **(111)**

"Ukraine needs these weapons now and not later, because later there may be no more Ukraine."

  Argument against **(112)**

"Germany should refrain from supplying heavy weapons, as this could provoke a third world war."

**###** Often, arguments contain **argument markers**, particularly, conclusion markers, such as "so", "therefore", "thus", "hence", or reason markers, such as "because", "for", "as", for the reason that", "since", etc.

**###** Sometimes, arguments do not formally contain all the necessary information to be coded yet it is clear from the context that they refer to the topic of arms deliveries and represent arguments from the linguistic point of view.

In such cases, we look at **two** sentences **before** and **after** the sentence containing the argument. If those sentences are enough to extract relevant meaning, we code 111 or 112. For example (two contextual sentences highlighted in gray):

"It is very dangerous to escalate in such a turbulent situation. And this especially applies to arms deliveries. They should be avoided by all means not to become a war party." (CODE: "112")

**###** Contextual sentences should **NOT** be coded unless they are themselves arguments or claims.

*Claims*

### Claim is a prescription/demand/viewpoint regarding (in)appropriateness of arms deliveries to Ukraine **without explanation**(s) of such position.

Claim for **(121)**

"Ukraine needs German weapons now and not later!"

Claim against **(122)**

"Ukraine should never receive German weapons!"

To understand whether a sentence is a claim, it may be useful to ask yourself the following question:
"Is X [the actor] explicitly for/against arms delivery?"

### Just as with arguments, we look at **two** sentences before and after a potential claim if there is not enough information within a single sentence. In the example below, the first sentence is a claim for arms deliveries. The second sentence is also a claim for arms deliveries even though without the context, it is not clear what "That" stands for. As we can extract that context from the previous sentence, we code 121.

"It is so that we strive to provide the weapons that are helpful and can be used well. (CODE 121). That is what we have done in the past, that is what we want to continue to do," Scholz said on Friday after a meeting with British Prime Minister Boris Johnson in London. (CODE 121).

### Sometimes, claims for/against arms deliveries are quoted directly, e.g., "We should deliver weapons to Ukraine." Yet more commonly, claims are reported or paraphrased, e.g., "X rejects the idea of sending arms." Thus, it is important not to miss **reported claims** on the topic of arms deliveries. They may contain such expressions as "X is hesitant about," "X calls for," "X supports the idea of," "X demands," "X requests," etc. Such sentences should be considered relevant because they imply an actor's stance regarding arms deliveries.

### In contrast, when **the fact of arms deliveries** rather than a **claim for/against arms deliveries** is reported, we do **NOT** code those sentences. Such sentences are mere descriptions of facts and do not provide clues regarding the stance of an actor on the topic of arms deliveries:

"Germany delivers 100 RPGs to Ukraine"

"Rheinmetall itself says that they have a whole series of Marder tanks that could be delivered in a very short time, he says to ZDF."

(The last example is about Rheinmetall's capability rather than desire because as a manufacturer, it does not have a political will).

### If there are doubts regarding whether a sentence is a claim or a description of a fact, we should pay attention to the context and format of the article. Generally, to avoid overinterpretation, a conservative approach is recommended in such cases.

### References to public opinion can be coded as arguments or as claims.

If reference to public opinion is used to justify a pro or contra position, it should be coded as an argument (111/112). For example:

"Weapons should not be delivered because the majority of Germans are against it." (CODE 112)

However, if a reference to public opinion is NOT used to justify a pro or contra position, it should be coded as 121/122. For example:

"According to the survey, 45 percent of respondents are in favor of arms deliveries." (CODE 121)

*Onion structure*

### Some arguments and claims may contain more than one actor, each with a stance on the topic of arms deliveries. We refer to such sentences as those with an **onion structure**. Usually, such sentences are structured as follows:

- Actor_1 has a stance X on the topic of arms deliveries;
- Actor_2 has a stance Y on the topic of arms deliveries, which is clear from how this actor reacts to the stance of Actor_1.

In such cases, **we always code a stance of Actor_1** (the actor who is directly commenting on the topic of arms deliveries rather than the one commenting on someone else's stance on the topic of arms deliveries). After that, we additionally apply code **888** to indicate that there is also the opposite stance present within the same sentence. We only apply 888 when the stances of Actor_1 and Actor_2 are different. For example:

Kira Rudyk is disappointed with the coalition's demands not to supply tanks. (CODE 122 / 888)

*Coding procedure*

### The coding of each document consists of three steps which should be implemented in the following order:

**(1)** we first read an article from the beginning and till the end without coding anything;

**(2)** after finishing this step, we read the article again and code 111, 112, 121, 122, and 888 where applicable, considering the two contextual sentences before and after each potentially relevant sentence;

**(3)** finally, we look through the coded segments one more time to double-check the adequacy of the coding.

## Appendix 3. Language models implementation

**UKP-SAM dataset**

The UKP-SAM dataset consists of 400 labeled documents with 25,209 sentences on eight controversial topics with the labels: pro/con/no argument. Since it has been shown that this leads to better results (Reimers et al., 2019), we adjust the input for this model in the format

$$\text{SAM Input} := [\text{topic}] </s> [\text{Input}]$$

where we use a new topic label "Waffenlieferung Ukraine", which is not in the SAM dataset, for fine-tuning on our dataset. We first train XLM-RoBERTa large on the SAM dataset with a learning rate of 5e-6 for two epochs, then we continue fine-tune it on our dataset analog to the baseline model.

**Pattern-exploiting training (PET)**

The naive pattern is defined as

PET Input 1 := Dies ist ein <mask> <mask> Waffenlieferungen an die Ukraine: [Input]

where <mask> specifies the token that the model should predict. For each class in our dataset, we formulate a sequence of two tokens that build the associated verbalizers to the naive pattern:

- argumentagainst: "argument gegen"
- argumentfor: "argument für"
- claimagainst: "claim gegen"
- claimfor: "claim für"
- nostance: "Satz ohne"

For comparison, we define a more elaborate pattern as

PET Input 2 := Dieser Satz <mask> <mask> <mask> Waffenlieferungen an die Ukraine: [Input]

with the following verbalizers:

- argumentagainst: "argumentiert gegen"
- argumentfor: "argumentiert für"
- claimagainst: "widerspricht"
- claimfor: "fordert"
- nostance: "ist neutral zu"

Here, four of the five verbalizers consist of three tokens each (note that token and word are not equivalent for the tokenization of XLM-RoBERTA inputs), only the verbalizer for claimfor consists of a single token. Due to this, PET Input 2 requires three masked tokens to fit the longest pattern. Besides the usage of PVPs, in PET the total loss is computed from a combination of the loss for correctly inserting the class labels and a general language model loss to prevent catastrophic forgetting. For this latter task, we use the entire (labeled) data set as an unlabeled data set.

**Adapter-based fine-tuning**

For the application of the standard fine-tuning for bottleneck adapters we use the basic Input. In analogy to fine-tuning of a transformer pretrained on SAM, we also train an adapter on the UKP-SAM dataset. After that, we freeze the SAM adapter and train stacked on top a task-based bottleneck adapter for our dataset. For this setting, we use the SAM Input.

**Adapter with PET-head**

For this, we only use the naive PVP (PET Input 1), because it is more intuitive to implement due to the fact that all verbalizers have the same number of tokens. The basis for this model is a standard bottleneck adapter. However, the standard classification head is replaced by our implementation of the PET head, which consists of a linear layer (with dimensions: 1024 to 1024), a GELU activation function, a layer normalization and a final linear layer (with dimensions: 1024 to 6 = verbalizer vocabulary size). From the last layer the logits for the respective masked tokens are extracted and reduced to the relevant tokens of the verbalizer. The sum of the logits for the corresponding first and second tokens of the verbalizers as first and second mask tokens result in the logits for the class labels. The loss is calculated using cross entropy.

For the near-domain fine-tuning in combination with our implementation of the PET-like classification head, we use the already pre-trained SAM adapter and use the following input for training:

$$\text{SAM+PET Input} := [\text{topic}] \text{</s>} [\text{PET Input 1}]$$

## Appendix 4. Error analysis

Table A1. Comparison of correctly and incorrectly classified arguments when stance was detected correctly. Argument markers and "trigger words" are highlighted in red. Where necessary, contextual information is provided in blue.

| actual = argument, predicted = claim (correct stance) | | actual = argument, predicted = argument (correct stance) | |
|---|---|---|---|
| Original | English | Original | English |
| Die Ukraine bittet schon seit langem um deutsche Hilfe auch durch Waffenlieferungen. Die Regierungsparteien SPD, Grüne und FDP haben sich in ihrem Koalitionsvertrag jedoch darauf festgelegt, keine Waffenlieferungen in Krisengebiete zuzulassen. | Ukraine has long been asking for German help, including through arms deliveries. However, the governing parties SPD, Greens and FDP have agreed in their coalition agreement not to allow arms deliveries to crisis areas. | Denn unsere Geschichte wäre, wenn überhaupt, das beste Argument FÜR Waffenlieferungen. | Because our history would be, if anything, the best argument FOR arms deliveries. |
| "Die deutsche Bundesregierung verfolgt seit vielen Jahren eine gleichgerichtete Strategie in dieser Frage. Und dazu gehört auch, dass wir keine letalen Waffen exportieren", sagte Scholz. | "For many years, the German federal government has pursued a like-minded strategy on this issue. And that also means that we don't export lethal weapons," said Scholz. | Angesichts der drohenden russischen Aggression ersucht die Ukraine eine möglichst zeitnahe Bearbeitung dieses Antrags [für Waffen]. | In view of the threat of Russian aggression, Ukraine requests that this application [for weapons] be processed as soon as possible. |
| "Die deutsche Bundesregierung verfolgt seit vielen Jahren eine gleichgerichtete Strategie in dieser Frage. Und dazu gehört auch, dass wir keine letalen Waffen exportieren", hatte Bundeskanzler Olaf Scholz (SPD) gesagt. | "For many years, the German federal government has pursued a like-minded strategy on this issue. And that also means that we don't export lethal weapons," said Federal Chancellor Olaf Scholz (SPD). | Dem Versuch Putins, im Donbass Fakten zu schaffen, müsse nun, so die zu erwartende Argumentation, damit begegnet werden, dass der Ukraine ausreichend Waffen zur Verfügung gestellt werden, um die besetzen Gebiete | According to the expected argument, Putin's attempt to create facts in the Donbas must now be countered by providing Ukraine with sufficient weapons to reconquer the occupied territories – and as quickly as possible. |
| | | zurückzuerobern – und das möglichst schnell. | |
| Jörg Kramer, Uelzen„Sofa-Pazifismus" wird Putin nicht aufhalten, und ich bin mittlerweile nicht zuletzt aufgrund des Aufrufs von Alice Schwarzer zur Überzeugung gelangt, dass der Bundestag mit der Lieferung von schweren Waffen in die Ukraine die richtige Entscheidung getroffen hat. | Jörg Kramer, Uelzen's "sofa pacifism" will not stop Putin, and I am now convinced, not least because of Alice Schwarzer's appeal, that the Bundestag made the right decision in supplying heavy weapons to Ukraine. | Der ukrainische Botschafter in Deutschland, Andrij Melnyk, sagte dem "Handelsblatt", der "Ernst der Lage" verlange von der Ampel-Regierung ein "sofortiges Umdenken" bei diesem Thema [die Unangemessenheit von Waffenlieferungen]. | The Ukrainian ambassador in Germany, Andriy Melnyk, told the "Handelsblatt" that the "seriousness of the situation" demanded an "immediate rethink" from the traffic light government on this issue [the inappropriateness of arms deliveries]. |
| Nach Hinweisen auf einen weiteren russischen Truppenaufwuchs an der Grenze zur Ukraine und den damit verbundenen Sorgen vor einem Angriff haben sich am Mittwoch mehrere deutsche Politiker dafür ausgesprochen, Waffen an die Ukraine zu liefern. | After indications of a further increase in Russian troops on the border with Ukraine and the associated fears of an attack, several German politicians spoke out in favor of supplying arms to Ukraine on Wednesday. | Es sei Konsens in der Regierung, dass Waffenlieferungen "aktuell nicht hilfreich" seien. | There is a consensus in the government that arms deliveries are "currently not helpful". |
| Angesichts des Beginns einer neuen russischen Großoffensive im Osten der Ukraine sieht sich Bundeskanzler Olaf Scholz (SPD) mit lauter werdenden Forderungen konfrontiert, der Lieferung schwerer Waffen zuzustimmen. | In view of the start of a new Russian offensive in eastern Ukraine, Chancellor Olaf Scholz (SPD) is faced with increasing calls for approval of the delivery of heavy weapons. | Bundeskanzler Olaf Scholz sagt, dass im Falle einer Konflikteskalation durch Russland alle Möglichkeiten geprüft würden. Aber Deutschland habe seit Jahren den Grundsatz, "keine letalen Waffen zu exportieren". Auch Baerbock verweist auf die restriktiven Exportrichtlinien Deutschlands. | German Chancellor Olaf Scholz says that in the event of an escalation of the conflict by Russia, all options would be examined. But Germany has for years had the principle "not to export lethal weapons". Baerbock also refers to Germany's restrictive export guidelines. |
| Auch der designierte CDU-Vorsitzende Friedrich Merz plädiert angesichts des russischen Truppenaufmarschs an der Grenze zur Ukraine dafür, Waffenlieferungen in Erwägung zu ziehen. | In view of the Russian troop deployment on the border with Ukraine, the designated CDU chairman Friedrich Merz also advocates considering arms deliveries. | Dort habe man Lieferungen in Krisengebiete ausgeschlossen, da "gehört die Ukraine dazu" [Krisengebiete]. | Deliveries to crisis areas have been ruled out there, because "Ukraine is part of it" [crisis areas]. |
| Die Vorsitzende des Verteidigungsausschusses, Marie-Agnes Strack-Zimmermann, seit jeher Verfechterin von verstärkten Waffenlieferungen, brauchte am Mittwoch gerade mal zwei Stunden für eine entsprechende Forderung auf Twitter: Die Teilmobilmachung, so die FDP-Politikerin, müsse als "Anlass dienen, unsere Bemühungen zur Unterstützung der Ukraine zu intensivieren". | The chairwoman of the Defense Committee, Marie-Agnes Strack-Zimmermann, who has always advocated increased arms deliveries, needed just two hours on Wednesday to make a corresponding request on Twitter: The partial mobilization, according to the FDP politician, must serve as "an occasion for our efforts to intensify support to Ukraine". | [Waffenlieferungen sind] mit den deutschen Rüstungsexportrichtlinien angeblich nicht vereinbar. | [Arms deliveries are] allegedly not compatible with the German arms export guidelines. |
| Kiew wünsche sich Kriegsschiffe zur Küstenverteidigung und Luftabwehrsysteme. | Kyiv would like warships for coastal defense and air defense systems. | Es sei "Konsens in der Bundesregierung", dass Waffenlieferungen in die Ukraine angesichts der zugespitzten Lage "aktuell nicht hilfreich" seien, sagte sie der "Welt am Sonntag". | There was a "consensus in the federal government" that arms deliveries to Ukraine were "currently not helpful" given the tense situation, she told the "Welt am Sonntag". |
| Die Ampel tritt für eine restriktive Rüstungspolitik ein, so steht es im | The coalition agreement states that the traffic light coalition is in favor of a | Sie halte es aber nicht für realistisch, mit solchen Lieferungen das militärische | However, she does not consider it realistic to reverse the military |
| Koalitionsvertrag. Waffenlieferungen in Krisengebiete sind darin nicht vorgesehen. | restrictive arms policy. It does not provide for arms deliveries to crisis regions. | Ungleichgewicht umzukehren. | imbalance with such supplies. |
| | | Das hat damit zu tun, dass die Bundesregierung keine Rüstungsgüter in Krisengebiete exportieren will, zudem will Berlin aufgrund seiner Vergangenheit nicht mit Waffenlieferungen in die ehemalige Sowjetunion vorpreschen. | This has to do with the fact that the federal government does not want to export armaments to crisis areas, and because of its past, Berlin does not want to rush forward with arms deliveries to the former Soviet Union. |

## Appendix 5. Few-shot settings

Table A2. Number of observations per label in the few-shot train set for the original dataset

| original FS setting | argumentfor | argumentagainst | claimfor | Claimagainst | nostance |
|---|---|---|---|---|---|
| 0.025 | 1 | 1 | 2 | 2 | 27 |
| 0.050 | 2 | 2 | 5 | 4 | 54 |
| 0.100 | 4 | 4 | 10 | 8 | 109 |
| 0.200 | 9 | 9 | 20 | 16 | 218 |
| 0.300 | 13 | 13 | 30 | 24 | 327 |
| 0.500 | 23 | 22 | 50 | 40 | 545 |
| 0.700 | 32 | 31 | 70 | 56 | 763 |
| 1.000 | 46 | 45 | 101 | 81 | 1091 |

Table A3. Number of observations per label in the few-shot train set for the onion-swapped dataset

| onion FS setting | argumentfor | argumentagainst | claimfor | claimagainst | nostance |
|---|---|---|---|---|---|
| 0.025 | 1 | 1 | 2 | 2 | 27 |
| 0.050 | 2 | 2 | 4 | 5 | 54 |
| 0.100 | 4 | 4 | 9 | 9 | 109 |
| 0.200 | 10 | 8 | 22 | 14 | 218 |
| 0.300 | 14 | 12 | 32 | 22 | 327 |
| 0.500 | 25 | 20 | 53 | 37 | 545 |
| 0.700 | 34 | 29 | 74 | 52 | 763 |
| 1.000 | 49 | 42 | 101 | 81 | 1091 |

## Appendix 6. Parameter robustness for fully fine-tuning

Figure A1. FSL performance evaluation for the fully fine-tuned model on 10 vs. 30 epochs

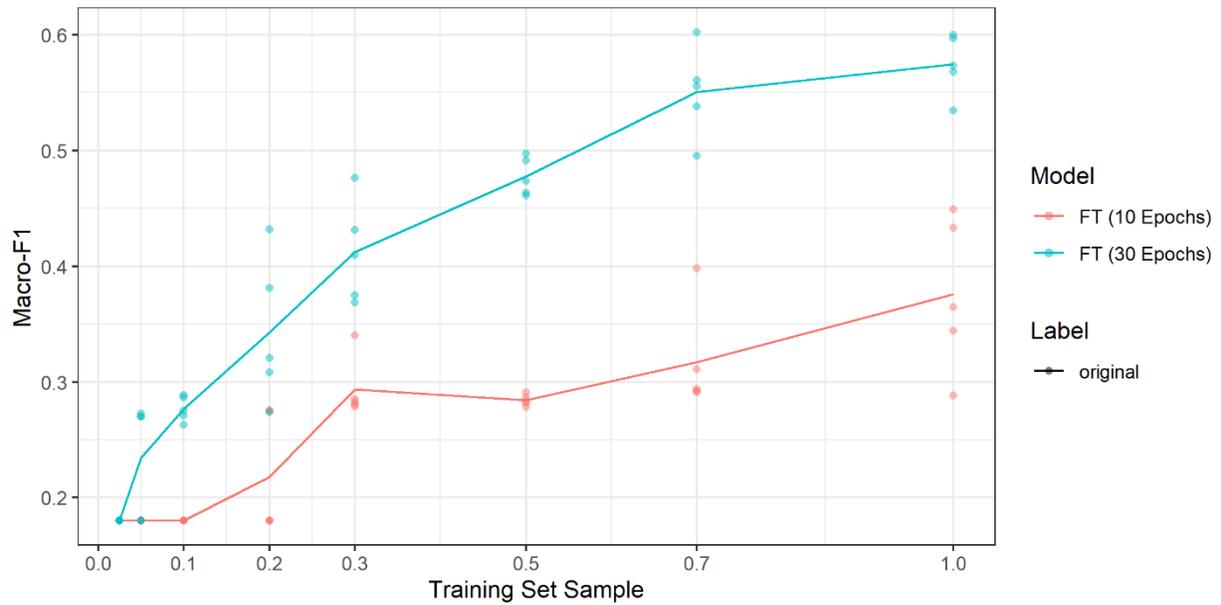

*Note.* Dots represent individual runs; the curve represent the mean values.